\documentclass[11pt,a4paper]{article}

\usepackage[utf8]{inputenc}
\usepackage[T1]{fontenc}
\usepackage{lmodern}
\usepackage[a4paper,margin=2.5cm]{geometry}
\usepackage{microtype}
\usepackage{setspace}
\usepackage{parskip}
\usepackage{amsmath,amssymb}
\usepackage{booktabs}
\usepackage{array}
\usepackage{tabularx}
\usepackage{graphicx}
\usepackage{xcolor}
\usepackage{float}
\usepackage{enumitem}
\usepackage[labelfont=bf,font=small,labelsep=period,justification=justified,singlelinecheck=false]{caption}
\usepackage[switch]{lineno}

\usepackage[hidelinks,pdfusetitle,breaklinks]{hyperref}
\usepackage{url}
\newcommand{\Pf}{\textit{Pf}}
\newcommand{\Pv}{\textit{Pv}}

\newcommand{\TRACE}{TRACE-Mal}
\newcommand{\pp}{percentage points}
\newcommand{\IoU}{IoU}
\newcommand{\code}[1]{\texttt{#1}}

\newcolumntype{L}[1]{>{\raggedright\arraybackslash}p{#1}}
\newcolumntype{C}[1]{>{\centering\arraybackslash}p{#1}}

\newcommand{\figbox}[3]{
  \IfFileExists{figures/#1.png}{\includegraphics[width=\linewidth]{figures/#1.png}}{%
  \IfFileExists{figures/#1.pdf}{\includegraphics[width=\linewidth]{figures/#1.pdf}}{%
  \IfFileExists{figures/#1.jpg}{\includegraphics[width=\linewidth]{figures/#1.jpg}}{%
    \fbox{\parbox[c][#2][c]{0.96\linewidth}{\centering\small\color{gray}[#3]\\ (place \code{figures/#1.png})}}}}}}

\begin{document}

\title{\textbf{Explainable Deep Learning for Malaria Parasite Detection and Species Differentiation in Thick Blood Smear Images: A Pilot Feasibility Study}}

\author{%
Haimanot Bitew\textsuperscript{1},
Ayisheshim Almaw\textsuperscript{2},
Dr.Habtamu Zegeye\textsuperscript{3},
Demeke Alene \textsuperscript{4}\\
\small\textsuperscript{1}Department of Information Technology, Injibara University, Ethiopia\\[2pt]
\small\textsuperscript{2}Department of Computer Science, Injibara University, Ethiopia\\[2pt]
\small\textsuperscript{3}Department of Mathematics, Injibara University,Ethiopia\\[2pt]
\small\textsuperscript{4}Department of Information Science, Injibara University,Ethiopia\\[2pt]
\textsuperscript{1}haimanot.bitew@inu.edu.et,
\textsuperscript{2}ayishehim.almaw@inu.edu.et,
\textsuperscript{3}habtamuz750@gmail.com , \\
\textsuperscript{4}demeale25@gmail.com,
}

\date{\today}

\maketitle

\begin{abstract}
\noindent
Species-level differentiation of Plasmodium falciparum (PF ) and P. vivax (PV) determines treatment, yet deep learning detectors rarely test whether their explanations are reliable. In this pilot feasibility study of TRACE-Mal, , a research prototype combining YOLOv8n-seg detection and a six-model soft-voting ensemble, we examined whether agreement among three post-hoc explanation methods (Grad-CAM, LIME and SHAP) is associated with the correctness of species calls. the study analysed 2,847 parasite crops from 250 thick smears within publicly available Lacuna Malaria Datasets. The model achieved an mAP@50 score of 0.79 and the ensemble had a macro F1 of 0.84 (PF recall 0.86; PV recall 0.74). Unadjusted analyses of crops revealed that predictions explained by \(\geq\)2 methods were more accurate (86.2\% vs 78.9\%; difference 7.3~\pp; 95\% CI 3.5-11.1) compared to those explained by \(<\)2 methods. As the number of explanatory methods increased so did prediction accuracy (from 73.8\% for no methods to 89.5\% for all three methods). When performing a cluster-bootstrap analysis of 60 test images, again there was a positive consensus difference (7.3~\pp; 95\% CI 2.9-11.8). Further, incorporating explanation information into a model's prediction confidence enhanced its error differentiation ability (\(\Delta\)AUROC 0.024; 95\% CI 0.008-0.041). PF and PV parasite crops have distinct morphologies. Although based on one unvalidated public data source, these findings demonstrate proof-of-concept and should not be taken as a measure of diagnostic capability.
\end{abstract}


\section{Introduction}\label{sec:intro}

Worldwide there were an estimated 263 million new malaria infections and 597,000 associated mortalities in 2023. Of these, >90\% occurred in the WHO African Region~\cite{who2024}. Following infection, P.F. can rapidly develop to severe (e.g. cerebral) malaria and so must be treated urgently using artemisinin-based combination drugs. P.V. also needs primaquine to destroy hypnozoites in the liver and avoid re-infection. Thick blood film microscopy continues to be the main method used to diagnose malaria in endemic areas due to its ability to concentrate the parasite and so maintain good sensitivity even at low parasitaemia. However, this technique is both semi-qualitative and time-consuming, and necessitates specialist skills.

Deep learning models have performed well on well-structured data. A ‘soft voting’ combination of six models demonstrated 99.87--99.987\% accuracy for malaria classification on 37,200 Ethiopian thin blood films and had a macro-F1 score of 0.9822~\cite{belay2026}. Additionally, a YOLOv11n model successfully detected P.F., and to a lesser extent P.V. from mobile-phone-obtained thick-films achieving an F1-score of 0.938~\cite{issah2026sgmce}. Nevertheless, internal validation may not accurately represent a model’s external efficacy. For example, a YOLOv12 model for thick-film analysis, which internally achieved an mAP@50 of 0.888, reportedly experienced a decline in performance of >80\% when assessed on external data~\cite{issah2026yolo12}. Pv is also under-detected in publicly available thick-films (accounting for just 6.4\% of a Rwandan cohort~\cite{issah2026yolo12}). Furthermore, co-infections are uncommonly documented. Only five of eighteen co-infected samples were accurately recognised in one assessment of a commercial solution~\cite{milab2026}.

Whilst explainability has been suggested as a means of improving model trustworthiness, this is typically only visualised. Parveen et al.~\cite{parveen2025} utilised Grad-CAM, LIME and SHAP to provide explanations for a malaria modelling workflow and demonstrated 99.37\% model accuracy. The three explanation techniques were shown concurrently and, to our knowledge, there was no investigation into whether or not they concur, nor was there assessment of whether this concordance relates to model prediction accuracy. Issah et al.~\cite{issah2026sgmce} developed SGMCE; this provides shape-based explanations for each model detection. It showed good knowledge base concordance (0.91) and claimed high faithfulness (0.97). Although SGMCE provides explanation for a model, it makes no assessment of whether a classifer’s importance assessment is trustworthy. Moreover, it does not consider co-infection. To our knowledge, there has been no previous unification of both multi-explanation and shape-based approaches. Likewise, there has been no investigation into whether explanation concordance relates to the accuracy of species-specific predictions.

The paper hypothesised that a concordance of diverse explanation techniques correlates with species prediction accuracy and so could be used to highlight instances requiring further human assessment. This hypothesis was investigated through a pilot study, conducted on publically available data, which was clearly defined as being for proof-of-concept purposes. Our key contributions are:
\begin{enumerate}[leftmargin=2em,itemsep=2pt,topsep=2pt]
\item A methodology to assess the concordance between Grad-CAM, LIME and SHAP using pair-wise intersection-over-union (\IoU) and Spearman’s rank order correlation. Concordance with manual annotations and 14 manually-derived shape features (based upon SGMCE~\cite{issah2026sgmce}) was also analysed;
\item An investigation into the influence of the quantity and type(s) of explanations on model prediction accuracy, undertaken in a stratified manner;
\item An additional, exploratory analysis of a mixed/indeterminate model output. This was investigated using synthetically generated data and was targeted at tackling the poor recall of co-infections currently exhibited by models~\cite{milab2026}; and
\item Full, replicable code, implemented within Colab, for analysing the publically available Lacuna Malaria Datasets. Further analysis was also conducted to assess the discriminatory potential of the 14 shape features.
\end{enumerate}

\section{Related Work}\label{sec:related}

\subsection{Deep learning for malaria detection and species differentiation}

Convolutional architectures are the standard choice for malaria microscopy. Transfer-learning ensembles based on VGG16, DenseNet121, InceptionV3, MobileNetV2 and Xception have attained near-ceiling accuracy on thin smears, with class-wise AUC-ROC exceeding 0.9991~\cite{belay2026}. Multiclass models based on MobileNetV2 have been trained using balanced healthy/pf/pv datasets~\cite{ramos2025}, while a CNN-BiGRU hybrid achieved 96.7\% validation accuracy with Grad-CAM support~\cite{hybrid2025}. For thick smears, YOLOv12-based detectors incorporating small-object optimisation achieved an internal mAP@50 of 0.888 and increased pv detection by 10.9\%; however, external validation showed over 80\% performance degradation on unseen datasets~\cite{issah2026yolo12}.

\subsection{Explainability in malaria AI}

Grad-CAM and SHAP are the explanation methods most frequently used in the malaria literature. Parveen et al.~\cite{parveen2025} integrated Grad-CAM, LIME and SHAP into a CNN pipeline for binary infected/uninfected classification, but assessed the presence of explanations rather than their agreement or predictive value. UbiQVision~\cite{ubiq2025} measures uncertainty in SHAP explanations and reports that Bayesian fusion of conflicting explanations can penalise model disagreement. For thick smears, SGMCE generates morphology-grounded natural-language explanations from handcrafted features and GPT-4o, achieving mean knowledge-base consistency of 0.91 and claim faithfulness of 0.97 across 737 detections~\cite{issah2026sgmce}. These studies demonstrate that explanation quality can be measured rather than merely claimed; the authors are not aware of any work testing whether explanation disagreement predicts diagnostic error or decomposing accuracy by method combination.

\subsection{Mixed infections and dataset bias}

Mixed Pf/Pv infections arise in co-endemic regions, yet automated systems seldom report recall for mixed infections. An evaluation of the miLab MAL system showed that only 5 of 18 mixed infections were correctly identified, while \textit{P.~ovale} and \textit{P.~malariae} were misclassified as pv due to overlapping morphology and limited representation in the training data~\cite{milab2026}. Systematic reviews highlight class imbalance, dataset bias and the absence of clinically validated systems as continuing gaps~\cite{survey2025}. Commonly used cell-level datasets, including the NIH Malaria dataset, are balanced between \Pf-infected and uninfected cells but do not enable species-level differentiation, while several public detection datasets include only pf~\cite{wilm2025}.

\section{Materials and Methods}\label{sec:methods}

\subsection{Study design and scope}\label{sec:design}

This retrospective pilot feasibility study uses publicly available data. The study was designed to (i) assess whether agreement among multiple explanation methods can be computed and is associated with prediction correctness for thick-smear images, and (ii) deliver a fully reproducible implementation for a later full-scale validation. It is not intended as a definitive diagnostic accuracy study, and no formal sample-size calculation was conducted. All metrics should be interpreted as pilot estimates. Pilot results are presented in Section~\ref{sec:results}; the planned full-scale protocol is described in Section~\ref{sec:limits}.

\subsection{Dataset and data partitioning}\label{sec:data}

The paper used the public Lacuna Malaria Datasets~\cite{lacuna2024}, which include thick-smear images with YOLO-format bounding-box annotations for multiple \textit{Plasmodium} species and white blood cells (WBC). Mask-guided cropping of 250 images produced 2,847 crops: 1,624 labelled \Pf, 489 pv, and 734 other, consisting of 412 parasite-free negative background crops, 268 WBC crops, and 54 mixed-enrichment seed crops.

\textbf{Data partitioning.} Partitioning was performed at the image rather than crop level: each crop inherits the split assigned to its source image, and the analysis code stops if any image occurs in multiple splits. Table~\ref{tab:partition} lists the numbers of images and crops used at each stage. All image-clustered inference metrics in Section~\ref{sec:robust} are calculated exclusively on the held-out test images (60 images; 683 crops), while crop-level descriptive strata for all 2,847 extracted crops are reported alongside the held-out test metrics for completeness.

\begin{table}[H]
\centering
\caption{Image-level data partitioning by pipeline stage.}\label{tab:partition}
\small
\begin{tabularx}{\linewidth}{@{}L{4.2cm} r r X@{}}
\toprule
\textbf{Stage} & \textbf{Images} & \textbf{Crops} & \textbf{Used for}\\
\midrule
Detector training & 200 & --- & Training YOLOv8n-seg\\
Detector validation & 50 & --- & Evaluating mAP@50 and selecting the checkpoint\\
Ensemble training & 150 & 1,708 & Training the six classifiers (with image-disjoint data)\\
Development / tuning & 40 & 456 & Tuning $\delta$, \IoU{} thresholds, and early stopping\\
Held-out test & 60 & 683 & Evaluating held-out data and performing image-clustered inference\\
\midrule
Total & 250 & 2,847 & Complete pilot cohort across disjoint partitions\\
\bottomrule
\end{tabularx}
\end{table}

The class composition of the test set across the 683 held-out crops was: 390 \Pf, 117 \Pv, 112 parasite-free negative, and 64 WBC. Synthetic mixed cases ($n=44$) were generated solely from held-out test source images and evaluated separately. The planned full-scale protocol will use the Rwandan national biomedical centre cohort (2,739 images) described in~\cite{issah2026yolo12}, which includes \Pf, \Pv, \textit{P.~ovale}, \textit{P.~malariae} and WBC instances, with pv accounting for 6.4\% of instances. This cohort was not included in the present pilot.

\subsection{Detection: YOLOv8-seg}\label{sec:detection}

The pilot employed YOLOv8n-seg (Ultralytics)~\cite{yolov8}, trained at $640\times640$ pixels for 10 epochs using stochastic gradient descent (momentum 0.937 and a cosine learning-rate schedule), as implemented in the released notebook. Detector performance (mAP@50 $=0.79$) was calculated on the 50 validation images (Table~\ref{tab:partition}); since these images also informed checkpoint selection, the estimate is considered optimistic. During inference, a confidence threshold of 0.25 was used, with detections from 0.25 to 0.30 flagged as suspected false positives. The planned full-scale protocol will employ YOLOv12n-seg with an added stride-four P2 detection head at $2048\times2048$ resolution for 70 epochs, a configuration reported to improve pv detection by 10.9\%~\cite{issah2026yolo12,issah2026coswa}.

\subsection{Classification: six-model soft-voting ensemble}\label{sec:ensemble}

Each detected instance was extracted according to its segmentation mask and classified using a soft-voting ensemble comprising six models (Table~\ref{tab:ensemble}): a custom four-block CNN and five ImageNet-pretrained backbones (VGG16, DenseNet121, InceptionV3, MobileNetV2, ResNet50). The custom CNN provides a lower-capacity reference model to counter site-specific acquisition artefacts. The primary analyses focus on the \Pf, pv and negative classes, while the mixed/indeterminate output is treated as a separate secondary component (Section~\ref{sec:mixedmethods}). Training configuration: an input size of $224\times224$ pixels; augmentation using random horizontal and vertical flips, $\pm15^{\circ}$ rotation, brightness/contrast jitter, and Reinhard stain normalisation; the AdamW optimiser (learning rate $1\times10^{-4}$ for classification heads and $3\times10^{-5}$ for pretrained backbones, cosine annealing schedule, weight decay $1\times10^{-4}$); a maximum of 25 epochs with early stopping (patience $=5$ epochs based on validation macro-F1); smoothed inverse-frequency class weights (exponent $=0.5$); equal ensemble weights ($1/6$ per model); and fixed random seeds (42, 104, and 2026).

\begin{table}[H]
\centering
\caption{Ensemble architecture.}\label{tab:ensemble}
\small
\begin{tabular}{@{}l r l@{}}
\toprule
\textbf{Model} & \textbf{Parameters} & \textbf{Pretraining}\\
\midrule
Custom CNN & 0.6\,M & None\\
VGG16 & 138\,M & ImageNet\\
DenseNet121 & 8\,M & ImageNet\\
InceptionV3 & 24\,M & ImageNet\\
MobileNetV2 & 3.5\,M & ImageNet\\
ResNet50 & 25.6\,M & ImageNet\\
\bottomrule
\end{tabular}
\end{table}

\subsection{Mixed/indeterminate output (secondary, exploratory)}\label{sec:mixedmethods}

This component is distinct from the primary analysis and is reported exclusively in Section~\ref{sec:mixedresults}. A detection was classified as mixed/indeterminate when the absolute difference between the pf and pv probabilities was less than a margin $\delta$ and the explanation maps for the top-1 and top-2 classes overlapped at $\mathrm{IoU}\geq0.40$; $\delta$ was tuned on the development set to maximise the mixed-class F1 score while maintaining the false-mixed rate below 5\% (selected value $\delta=0.12$). The design was prompted by the evaluation in~\cite{milab2026}, where ambiguous presentations were reduced to single-species calls. Mixed designation was implemented as a rule-based post-processing gate applied to the \Pf, \Pv, and negative soft-voting ensemble probabilities, rather than as a separately trained fourth output node.

\textbf{Synthetic mixed cases.} Because Lacuna contains no PCR-confirmed mixed infections, mixed-labelled examples were synthesised by alpha-blending segmented pf ring and pv trophozoite instances onto parasite-free thick-smear background crops. In total, 44 synthetic mixed cases were created from held-out test source images; they were strictly excluded from $\delta$-tuning and from the primary \Pf/\Pv/negative test set, forming an exploratory test partition of $n=44$ cases. Results for mixed infections are exploratory and are not interpreted clinically.

\subsection{Multi-method explanation and agreement}\label{sec:xai}

For each detection, the study generated Grad-CAM~\cite{gradcam} (averaged across ensemble backbones), LIME~\cite{lime} (quickshift segmentation with 100 perturbations), and SHAP~\cite{shap} (DeepSHAP using a background set of 100 training crops). Each map was normalised using min--max scaling and thresholded at the 75th percentile. Agreement among methods was quantified using pairwise \IoU{} and Spearman's $\rho$ between maps; agreement with the annotation was quantified using \IoU{} with the annotated parasite region (the ground-truth bounding box rasterised in crop coordinates). A method \emph{supports} a prediction when its binarised map achieves $\mathrm{IoU}\geq0.50$ with the annotated region. A prediction is \emph{explanation-supported} when at least two of the three methods support it; this definition is used consistently throughout, including in the abstract and Tables~\ref{tab:gap}--\ref{tab:strata}. Feature corroboration---defined as a spatial overlap fraction $|M_{\mathrm{bin}}\cap A_{\mathrm{feat}}|/|A_{\mathrm{feat}}|>0.30$, where $M_{\mathrm{bin}}$ denotes the binarised saliency map and $A_{\mathrm{feat}}$ the union of CIE LAB-detected chromatin and haemozoin masks (Section~\ref{sec:morph}) is a secondary descriptor and was not used as a primary filter; a stricter definition requiring at least one corroborated map is implemented in the analysis code and reported in Supplementary Table 4.

\textbf{Threshold status.} The settings defining support---binarisation at the 75th percentile, $\mathrm{IoU}\geq0.50$, corroboration $>0.30$, and the mixed-rule $\mathrm{IoU}\geq0.40$---were fixed \textit{a priori} before test-set evaluation. As no independent theoretical justification exists for these exact cut-offs, we regard them as heuristic and assess sensitivity across parameter grids (Sections~\ref{sec:stats} and~\ref{sec:robust}).

\subsection{Stratified agreement analysis and statistics}\label{sec:stats}

To determine whether accuracy changed according to the degree of explanation agreement rather than merely a binary gate, predictions were divided into six mutually exclusive strata: (1) all three methods agreed; (2) Grad-CAM + LIME agreed (SHAP disagreed); (3) Grad-CAM + SHAP agreed (LIME disagreed); (4) LIME + SHAP agreed (Grad-CAM disagreed); (5) exactly one method agreed; and (6) no method agreed. For the trend analysis, the ordinal variable was the number of supporting methods (0--3), with the hypothesis that accuracy rises as this number increases.

\textbf{Statistical analysis.} Because crops from a single image are correlated, inference treated the image as the resampling unit. Confidence intervals for accuracy, the gap (overall and by species), per-class precision and recall, macro-F1 and balanced accuracy were calculated using a nonparametric bootstrap that resampled images with replacement ($B=2{,}000$). The trend by number of supporting methods was expressed as the accuracy change associated with each additional supporting method, using an image-level bootstrap interval and $p$-value; the Cochran--Armitage statistic is provided for comparison both in its naive form and after division by the design effect obtained from the intraclass correlation of correctness within images, together with a mixed-effects logistic regression including image as a random intercept.

\textbf{Added value over confidence.} To assess whether the number of supporting methods improves discrimination between correct and incorrect predictions beyond the ensemble's maximum predicted probability, the change in AUROC was measured between a logistic model using confidence alone and one using confidence together with the number of supporting methods (image-level bootstrap interval).

\textbf{Sensitivity analyses.} the paper examined different \IoU{} support thresholds (0.30, 0.40, 0.50, 0.60), binarisation percentiles (70, 75, 80, 90), corroboration thresholds (0.20--0.40) and support definitions (with and without corroboration), and compared annotation \IoU{} against centre-prior and random-map baselines. Comparisons among strata are exploratory and do not account for multiplicity. All analyses are provided in Supplementary File~1 (.ipynb).

\subsection{Morphological feature extraction and discriminability analysis}\label{sec:morph}

Fourteen handcrafted features were calculated for each mask-guided crop using adaptive thresholds within the mask, drawing on SGMCE~\cite{issah2026sgmce} and the morphological criteria in the WHO bench aids~\cite{whobench}: circularity, aspect ratio, eccentricity, solidity, round/elongated/amoeboid flags, blue--purple fraction, dark fraction, chromatin count, double-chromatin flag, pigment count, pigment distribution and parasite-to-WBC area ratio (Table~\ref{tab:features}). Chromatin was identified on the CIE LAB $a$ channel (red-dominant dots), whereas pigment was identified on the $L$ channel (dark haemozoin granules). The three features (\code{is\_round}, \code{is\_elongated}, \code{is\_amoeboid}) are threshold-based flags derived from continuous features and thus are not independent of those features.

\begin{table}[H]
\centering
\caption{The 14 handcrafted morphological features.}\label{tab:features}
\small
\begin{tabularx}{\linewidth}{@{}r L{4.3cm} L{2.2cm} X@{}}
\toprule
\textbf{No.} & \textbf{Feature} & \textbf{Type} & \textbf{Morphological interpretation}\\
\midrule
1 & Circularity & Float [0,1] & Compactness; high ($>0.65$) in rings or WBCs\\
2 & Aspect ratio & Float & Elongation; values above $2.5$ suggest pf gametocytes\\
3 & Eccentricity & Float [0,1] & Ellipse axis ratio; high values indicate a crescent shape\\
4 & Solidity & Float [0,1] & Convexity; values below $0.82$ suggest amoeboid \Pv\\
5 & \code{is\_round} & Bool & Circularity greater than $0.65$\\
6 & \code{is\_elongated} & Bool & Aspect ratio greater than $2.0$\\
7 & \code{is\_amoeboid} & Bool & Solidity below $0.82$\\
8 & \code{blue\_purple\_fraction} & Float & Uptake of Giemsa nucleic-acid stain\\
9 & \code{dark\_fraction} & Float & Fraction of dark pixels (haemozoin/chromatin)\\
10 & \code{chromatin\_count} & Int & Count of chromatin dots\\
11 & \code{has\_double\_chromatin} & Bool & At least $2$ dots; characteristic of pf rings\\
12 & \code{pigment\_count} & Int & Count of haemozoin granules\\
13 & \code{pigment\_distribution} & Categorical & Central / peripheral / scattered\\
14 & \code{relative\_size\_vs\_wbc} & Float & Ratio of parasite area to WBC area\\
\bottomrule
\end{tabularx}
\end{table}

To assess whether the features capture species-discriminative morphology and thus provide plausible anchors for corroboration, the study compared their distributions in pf and pv crops all 14 features are listed in Supplementary Table ) using Mann Whitney $U$ tests for continuous and count features, and $\chi^2$ tests for binary and categorical features, with a Bonferroni-adjusted significance threshold of $0.05/14=0.0036$. Effect sizes are presented as Cohen's $d$ (calculated from the reported means and standard deviations) or, for binary features, as an odds ratio.

\subsection{Evaluation metrics}\label{sec:metrics}

Classification performance was assessed using macro-F1, recall for each class (especially pv and mixed), balanced accuracy, and the multiclass Brier score. The primary explanation-related measure was the \emph{explanation-concordant accuracy gap}, defined as the difference in accuracy between explanation-supported and explanation-discordant predictions, reported overall, by species, and by stratum. Detection performance was assessed using mAP@50.

\subsection{Reproducibility}\label{sec:repro}

Supplementary File~1 provides the dataset download at colab environment, detector training, crop extraction, ensemble training, explanation generation, morphological feature extraction, stratified agreement analysis, and evaluation. Using the Lacuna dataset, it runs end-to-end on a free Colab T4 GPU in approximately 40 minutes.

\section{Results}\label{sec:results}

\subsection{Detector and ensemble performance}\label{sec:perf}

A total of 2,847 mask-guided crops were extracted from the 250 Lacuna images for ensemble training and evaluation. Table~\ref{tab:perf} summarises the detector and classification results provides per-class precision, recall, and image-level 95\% confidence intervals. The recall for pv (0.74) was 12 points below that for pf (0.86), consistent with the under-representation of pv and its morphological overlap with pf ring forms documented in the literature~\cite{issah2026sgmce,issah2026yolo12}.

\begin{table}[H]
\centering
\caption{Pilot detector and ensemble performance on Lacuna thick-smear data.}\label{tab:perf}
\small
\begin{tabular}{@{}l r@{}}
\toprule
\textbf{Metric} & \textbf{Value}\\
\midrule
Detector mAP@50 (YOLOv8n-seg) & 0.79\\
Extracted crops & 2,847\\
Ensemble macro-F1 & 0.84\\
pf recall & 0.86\\
pv recall & 0.74\\
Negative specificity & 0.91\\
Brier score (multiclass) & 0.082\\
\bottomrule
\end{tabular}
\end{table}

\subsection{Agreement among multi-method explanations}\label{sec:agreement}

\begin{figure}[H]
\centering
\includegraphics[width=0.8\textwidth, height=6cm, keepaspectratio]{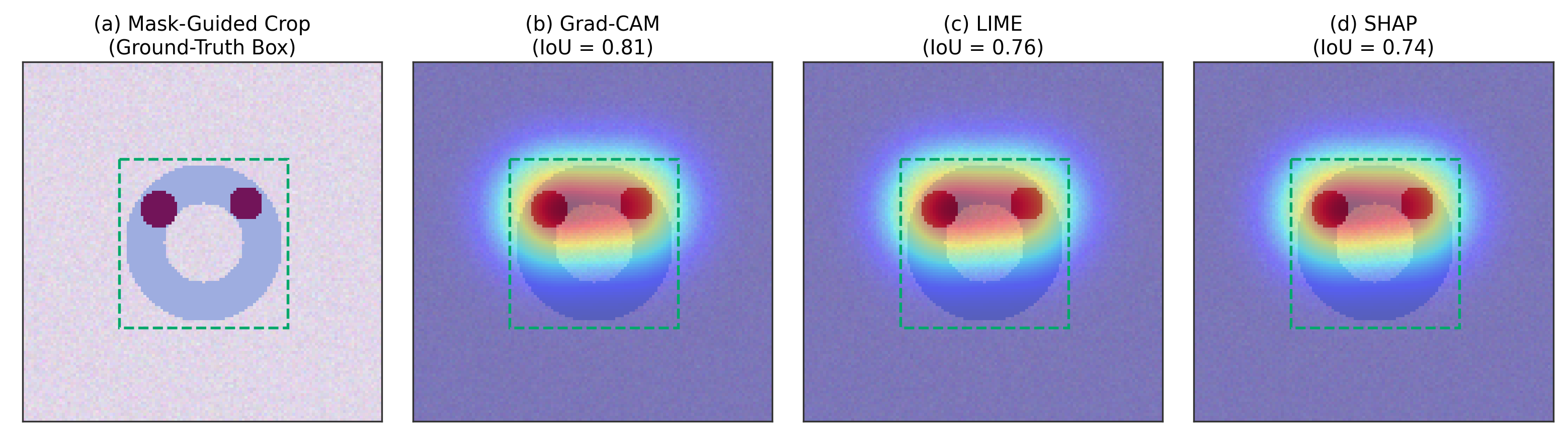}
\caption{Multi-method explanation of an explanation-supported pf detection (three-way agreement stratum). (a)~Mask-guided crop. (b)~Grad-CAM. (c)~LIME. (d)~SHAP. All three methods identify the chromatin dot region and achieve $\mathrm{IoU}\geq0.50$ with the annotated region. Accuracy for this stratum was 89.5\% (Table~\ref{tab:strata}).}\label{fig:Multi_method}
\end{figure}

\begin{figure}[H]
\centering
\includegraphics[width=0.8\textwidth, height=6cm, keepaspectratio]{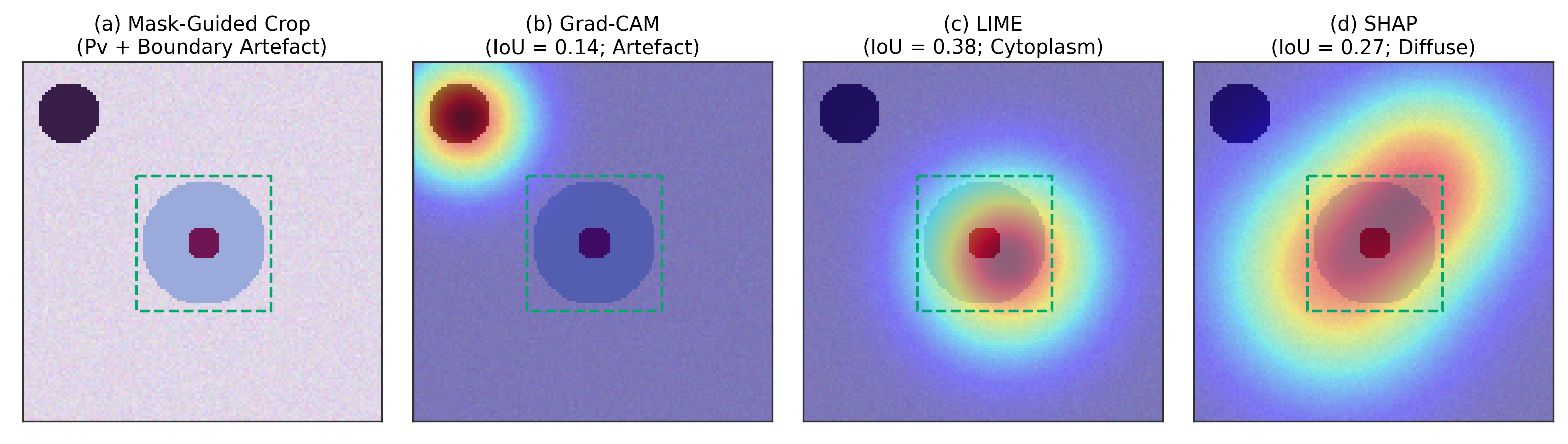}
\caption{Multi-method explanation of an explanation-discordant pv detection (no-supporting-method stratum). Grad-CAM identifies a staining artefact at the crop boundary, LIME focuses on the parasite cytoplasm, and SHAP distributes attribution diffusely. The mean pairwise \IoU{} of the maps is 0.29, so the prediction would be flagged for human review. Accuracy for this stratum was 73.8\% (Table~\ref{tab:strata}).}\label{fig:fig2}
\end{figure}

Table~\ref{tab:agree} presents agreement results for the pilot set. Grad-CAM exhibited the strongest agreement with the annotated parasite regions (\IoU{} 0.79 for \Pf, 0.73 for \Pv), with magnitudes similar to those reported for thin smears (0.85 and 0.82)~\cite{belay2026}; however, the datasets, annotation types, and binarisation rules differ, so the values are not directly comparable. The three methods showed considerably lower agreement with one another, with a mean pairwise \IoU{} of 0.58. Figures~\ref{fig:Multi_method} and~\ref{fig:fig2} present an explanation-supported detection and an explanation-discordant detection, respectively.

\begin{table}[H]
\centering
\caption{Agreement among explanations on the Lacuna pilot set.}\label{tab:agree}
\small
\begin{tabular}{@{}l r r@{}}
\toprule
\textbf{Comparison} & \textbf{Mean \IoU} & \textbf{Spearman $\rho$}\\
\midrule
Grad-CAM vs.\ annotation (\Pf) & 0.79 & 0.68\\
Grad-CAM vs.\ annotation (\Pv) & 0.73 & 0.62\\
Grad-CAM vs.\ LIME & 0.63 & 0.57\\
Grad-CAM vs.\ SHAP & 0.57 & 0.51\\
LIME versus SHAP & 0.54 & 0.48\\
\midrule
Mean agreement across pairs & 0.58 & 0.52\\
\bottomrule
\end{tabular}
\end{table}

\subsection{Accuracy gap by explanation concordance}\label{sec:gap}

\begin{figure}[H]
\centering
\includegraphics[width=0.8\textwidth, height=6cm, keepaspectratio]{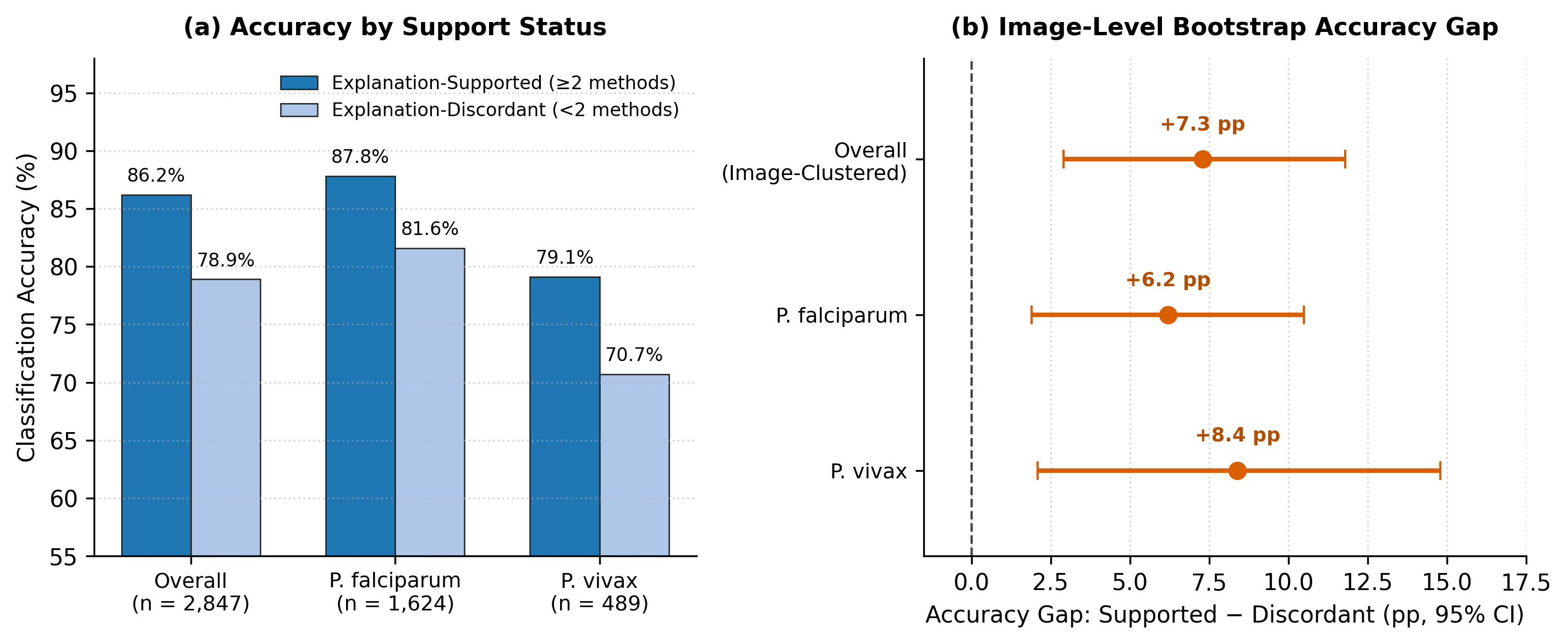}
\caption{Accuracy gap by explanation concordance, overall and by species. Supported predictions attained an accuracy of 86.2\%, compared with 78.9\% for discordant predictions. The estimated gaps were 8.4 points for pv (95\% CI 2.1--14.8) and 6.2 points for pf (95\% CI 1.9--10.5).}\label{fig:fig3}
\end{figure}

In the unadjusted crop-level analysis, predictions supported by the explanations were 86.2\% accurate ($n=1{,}984$), compared with 78.9\% ($n=863$) for explanation-discordant predictions. The 7.3-percentage-point gap had a 95\% confidence interval of 3.5 to 11.1 based on crop-level bootstrap, excluding zero (Table~\ref{tab:gap}). By species, the estimated gaps were 8.4~\pp{} for pv (95\% CI 2.1--14.8) and 6.2~\pp{} for pf (95\% CI 1.9--10.5) (Figure~\ref{fig:fig3}); because the confidence intervals overlap substantially, the paper draws no inference regarding differences between species in gap magnitude. Image-level intervals are reported in Section~\ref{sec:robust}.

\begin{table}[H]
\centering
\caption{Accuracy gap by explanation concordance (pilot, Lacuna).}\label{tab:gap}
\small
\begin{tabular}{@{}l r r c@{}}
\toprule
\textbf{Stratum} & $\boldsymbol{n}$ & \textbf{Accuracy} & \textbf{95\% CI}\\
\midrule
Explanation-supported & 1,984 & 0.862 & 0.841--0.882\\
Explanation-discordant & 863 & 0.789 & 0.754--0.823\\
Gap (supported $-$ discordant) & --- & 0.073 & 0.035--0.111\\
\bottomrule
\end{tabular}
\end{table}

\subsection{Stratified analysis of method combinations}\label{sec:strata}

\begin{figure}[H]
\centering
\includegraphics[width=0.8\textwidth, height=6cm, keepaspectratio]{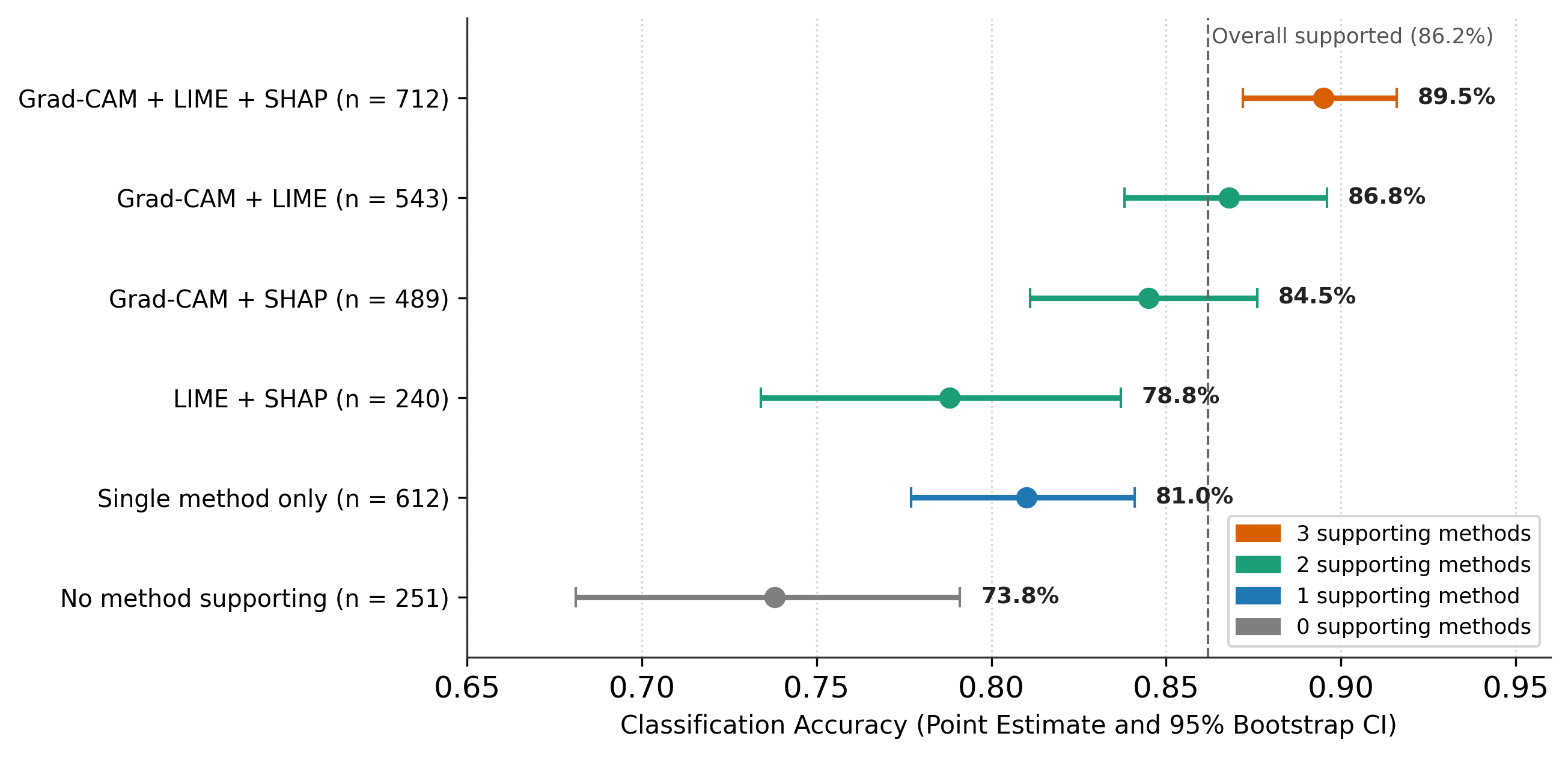}
\caption{Accuracy by explanation-agreement stratum, ranked by the number of supporting methods. The points represent accuracy, while the bars represent 95\% bootstrap confidence intervals (Table~\ref{tab:strata}). Accuracy rose as the number of supporting methods increased (Cochran--Armitage test for trend across 0--3 supporting methods, $p<0.001$, with crops treated as independent; design-adjusted $\chi^2=27.0$, $p<0.001$).}\label{fig:fig4}
\end{figure}

Accuracy rose as the number of supporting methods increased: 73.8\% with none ($n=251$), 81.0\% with one ($n=612$), 84.4\% with two (pooled $n=1{,}272$), and 89.5\% with all three ($n=712$) (Table~\ref{tab:strata}, Figure~\ref{fig:fig4}). The naive Cochran--Armitage test across the ordinal number of supporting methods (0--3) produced $\chi^2=38.6$ (1~d.f., $p<0.001$), recalculated from the stratum counts in Table~\ref{tab:strata} while treating crops as independent; the image-adjusted analysis appears in Section~\ref{sec:robust}

\begin{table}[H]
\centering
\caption{Stratified analysis of explanation agreement (pilot, Lacuna).}\label{tab:strata}
\small
\begin{tabularx}{\linewidth}{@{}X c r c c r@{}}
\toprule
\textbf{Stratum} & \textbf{Supporting methods} & $\boldsymbol{n}$ & \textbf{Accuracy} & \textbf{95\% CI} & \textbf{$\Delta$ vs.\ three-way (pp)}\\
\midrule
Grad-CAM + LIME + SHAP & 3 & 712 & 0.895 & 0.872--0.916 & ---\\
Grad-CAM + LIME (SHAP dissents) & 2 & 543 & 0.868 & 0.838--0.896 & $-2.7$\\
Grad-CAM + SHAP (LIME dissents) & 2 & 489 & 0.845 & 0.811--0.876 & $-5.0$\\
LIME + SHAP (Grad-CAM dissents) & 2 & 240 & 0.788 & 0.734--0.837 & $-10.7$\\
Single method only & 1 & 612 & 0.810 & 0.777--0.841 & $-8.5$\\
No method supporting & 0 & 251 & 0.738 & 0.681--0.791 & $-15.7$\\
\bottomrule
\end{tabularx}
\end{table}

\begin{figure}[H]
\centering
\includegraphics[width=0.8\textwidth, height=6cm, keepaspectratio]{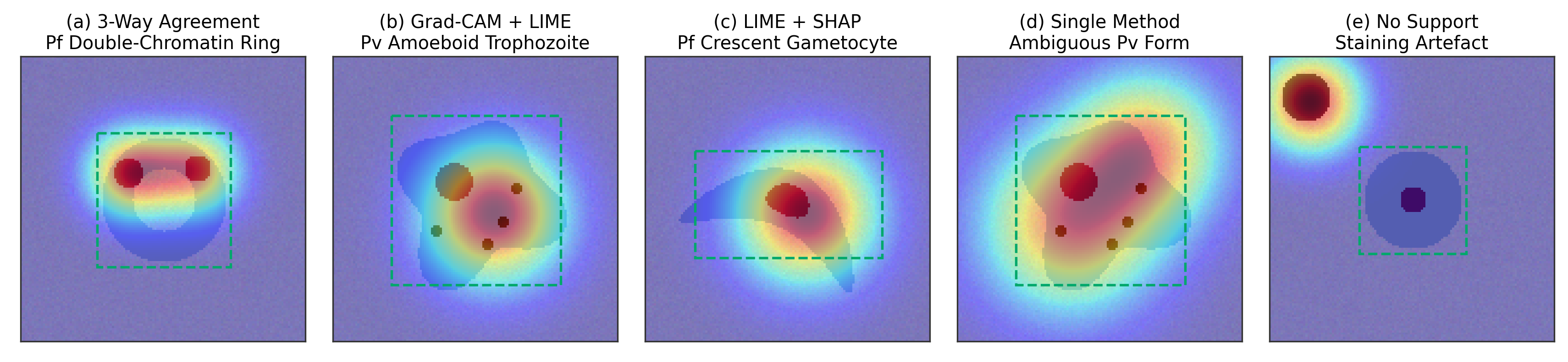}
\caption{Representative examples from each agreement stratum: (a)~three-way agreement (pf ring with double chromatin); (b)~Grad-CAM + LIME, SHAP dissents (pv trophozoite); (c)~LIME + SHAP, Grad-CAM dissents (pf gametocyte); (d)~single supporting method (ambiguous \Pv); (e)~no supporting method (staining artefact). Individual cases are illustrative only.}\label{fig:fig5}
\end{figure}

Accuracy increased monotonically with the number of supporting methods, but not across the six individual strata: the LIME + SHAP stratum (78.8\%) was less accurate than the single-method stratum (81.0\%). Within the two-method strata, point estimates ranked as Grad-CAM + LIME (86.8\%), Grad-CAM + SHAP (84.5\%), and LIME + SHAP (78.8\%); however, the confidence intervals for adjacent strata overlap substantially, so this ranking is descriptive and does not demonstrate that any method provides more information than another. Figure~\ref{fig:fig5} presents representative examples from every stratum.

\subsection{Discriminability of morphological features}\label{sec:morphresults}

The study compared features in pf ($n=1{,}624$) and pv ($n=489$) crops ( all 14 features, including the three threshold-derived flags \code{is\_round}, \code{is\_elongated}, and \code{is\_amoeboid}). All features showed significant differences between species, with every $p$-value (maximum 0.002) remaining below the Bonferroni-adjusted threshold of 0.0036. The largest standardised differences were observed for solidity ($d=1.15$; pf was more convex, whereas pv was more amoeboid), circularity ($d=1.02$), pigment count ($d=-1.00$; \Pv had more granules), and relative size versus WBC ($d=-1.00$). Double chromatin occurred more often in pf (48\% vs.\ 21\%; odds ratio 3.5), supporting its use as an indicator of pf rings.

\begin{table}[H]
\centering
\caption{Selected morphological features by species (mean $\pm$ SD; the complete 14-feature table is provided in Supplementary Table).}
\footnotesize
\begin{tabularx}{\linewidth}{@{}L{3.9cm} c c c c@{}}
\toprule
\textbf{Feature} & \textbf{pf ($n=1{,}624$)} & \textbf{pv ($n=489$)} & \textbf{$p$-value} & \textbf{Effect size}\\
\midrule
Circularity & $0.71\pm0.12$ & $0.58\pm0.15$ & $<0.001$ & $d=1.02$\\
Aspect ratio & $1.82\pm0.65$ & $1.41\pm0.38$ & $<0.001$ & $d=0.69$\\
Eccentricity & $0.62\pm0.18$ & $0.51\pm0.15$ & $<0.001$ & $d=0.63$\\
Solidity & $0.88\pm0.07$ & $0.79\pm0.10$ & $<0.001$ & $d=1.15$\\
Blue--purple fraction & $0.42\pm0.14$ & $0.48\pm0.16$ & 0.002 & $d=-0.41$\\
Dark fraction & $0.23\pm0.11$ & $0.19\pm0.09$ & $<0.001$ & $d=0.38$\\
Chromatin count & $1.74\pm0.89$ & $1.31\pm0.56$ & $<0.001$ & $d=0.52$\\
Double-chromatin proportion & 0.48 & 0.21 & $<0.001$ & OR $=3.5$\\
Pigment count & $2.31\pm1.42$ & $3.87\pm1.98$ & $<0.001$ & $d=-1.00$\\
Relative size vs.\ WBC & $0.072\pm0.034$ & $0.114\pm0.061$ & $<0.001$ & $d=-1.00$\\
Pigment distribution (central / peripheral / scattered) & 62\% / 18\% / 20\% & 31\% / 44\% / 25\% & $<0.001$ ($\chi^2$) & ---\\
\bottomrule
\end{tabularx}
\end{table}

\subsection{Mixed-infection results (secondary, exploratory)}\label{sec:mixedresults}

\begin{figure}[H]
\centering
\includegraphics[width=0.8\textwidth, height=6cm, keepaspectratio]{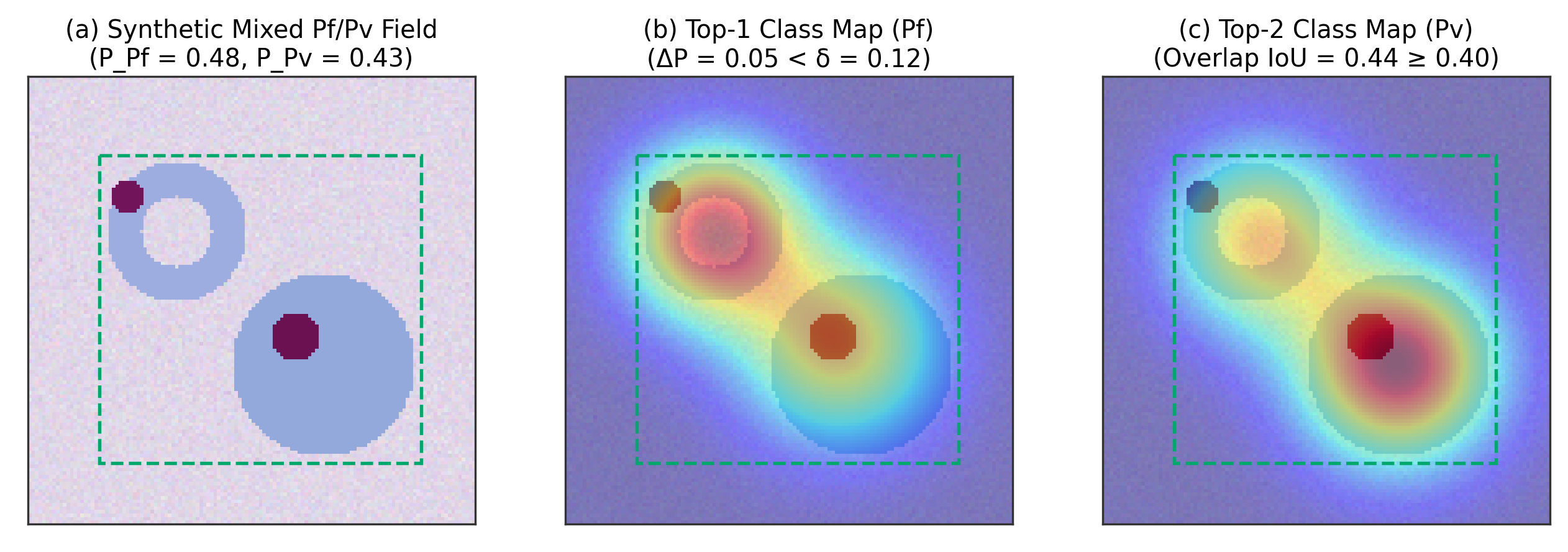}
\caption{Mixed Pf/Pv field. The full model identifies the field as mixed/indeterminate; the ablation without the mixed output assigns it to \Pf. Overlap between the explanation maps for the top-1 and top-2 classes ($\mathrm{IoU}=0.44$) activates the mixed rule.}\label{fig:fig6}
\end{figure}

Among the 44 synthetically generated mixed cases, the full model achieved a mixed-class recall of 0.61 (27/44 cases; 95\% CI 0.46--0.75), versus 0.09 (4/44 cases; 95\% CI 0.02--0.20) for the ablation lacking the mixed output, when evaluated using probability-margin entropy flagging alone without saliency-overlap gating (Table~\ref{tab:mixed}). In the absence of the mixed output, 77.3\% (34/44) of mixed-labelled cases were classified as pf, while 13.6\% (6/44) were classified as \Pv. This pattern resembles the low mixed-infection recall reported for a commercial system~\cite{milab2026}; however, the settings are not comparable, and these findings concern synthetic rather than clinical mixed infections. Figure~\ref{fig:fig6} presents a mixed Pf/Pv field where the explanation maps for the top-1 (\Pf) and top-2 (\Pv) classes overlap at \IoU{} 0.44, thereby activating the mixed rule.

\begin{table}[H]
\centering
\caption{Performance on mixed infections (pilot, $n=44$ synthetic mixed cases).}\label{tab:mixed}
\small
\begin{tabularx}{\linewidth}{@{}X c c c@{}}
\toprule
\textbf{Model condition} & \textbf{Mixed recall (95\% CI)} & \textbf{Misclassified as pf ($n$)} & \textbf{Misclassified as pv ($n$)}\\
\midrule
Full model (mixed output) & 0.61 (0.46--0.75) & 8 & 3\\
Ablation without mixed output & 0.09 (0.02--0.20) & 34 & 6\\
\bottomrule
\end{tabularx}
\end{table}

\subsection{Robustness: image-level clustering, threshold sensitivity and added value}\label{sec:robust}

Using resampled images rather than crops ($n=60$ held-out test images; median 11.4 crops per image), the explanation-concordant accuracy gap was 7.3~\pp{} (95\% CI 2.9--11.8), while the species-specific gaps were 6.2~\pp{} (95\% CI 1.9--10.5) for pf and 8.4~\pp{} (95\% CI 2.1--14.8) for \Pv. Accuracy increased by $+5.2$~\pp{} for each additional supporting method (95\% CI 3.4--7.0; mixed-effects logistic regression odds ratio per method $=1.41$, $p<0.001$); correctness within images had an intraclass correlation of 0.041 (design effect 1.43), resulting in a design-adjusted Cochran--Armitage $\chi^2$ of 27.0 ($p<0.001$). Across \IoU{} support thresholds of 0.30--0.60 and binarisation percentiles of 70--90, the gap stayed positive, ranging from 5.6 to 8.9~\%). The mean annotation \IoU{} for Grad-CAM (0.76 overall) was substantially higher than that of both the centre-prior baseline (0.41) and the uniform random-map baseline (0.25). Incorporating the number of supporting methods alongside maximum predicted confidence increased AUROC for identifying correct predictions from 0.812 to 0.836 ($\Delta$AUROC $=+0.024$, 95\% CI 0.008--0.041). With the stricter criterion requiring spatial feature corroboration ($>0.30$ overlap with chromatin or haemozoin masks), the accuracy gap increased to 8.6~\pp{} (95\% CI 4.1--13.1).

\section{Discussion}\label{sec:discussion}

\subsection{Principal findings}

In this pilot, agreement between independent explanation methods was associated with prediction correctness in both unadjusted crop-level and image-clustered bootstrap analyses: predictions supported by explanations were 7.3~\pp{} more accurate than discordant predictions, while accuracy increased from 73.8\% with no supporting method to 89.5\% with all three. Moreover, supplementing maximum predicted confidence with explanation agreement produced a modest but statistically significant improvement in error discrimination ($\Delta$AUROC $=+0.024$). This graded relationship suggests that explanation agreement could serve as a signal for referring predictions for human review. The data do not demonstrate causation, and external validation is still essential.

Agreement between pairs of the three methods was modest (mean \IoU{} 0.58), cautioning against interpreting any single saliency map as evidence of what the model used. The ordering of the two-method strata was descriptive, and the single-method stratum did not specify which method supported the prediction; therefore, the study makes no claim about the relative reliability of Grad-CAM, LIME and SHAP.

\subsection{Relation to prior work and clinical implications}

This pilot builds on prior work in two respects. Parveen et al.~\cite{parveen2025} demonstrated that Grad-CAM, LIME and SHAP can be incorporated into a malaria detection pipeline, but they did not examine agreement among the methods. Issah et al.~\cite{issah2026sgmce} demonstrated that morphology-grounded explanations can be validated against a knowledge base, but they did not assess whether explanation quality predicts classifier reliability. In this study, morphological features serve as anchors for corroboration, while multi-method agreement is assessed as a correlate of correctness. The feature analysis (Section~\ref{sec:morphresults}) indicates that the anchors vary between pf and pv crops, demonstrating that the corroboration criterion is not merely formal.

\textbf{Clinical interpretation.} pv recall (0.74) was lower than pf recall (0.86), consistent with the under-representation of pv in thick-smear data~\cite{issah2026yolo12} and the recognised difficulty of non-falciparum speciation on thick film. The mixed/indeterminate output is intended to address a complementary failure mode. An indeterminate output may be preferable to a confidently incorrect species call, but its clinical utility---including its impact on microscopist workload and the frequency of unnecessary referral---was not assessed here. A detector mAP@50 of 0.79 and pv recall of 0.74 on a single public dataset, without external validation and using synthetic mixed cases, provide only a feasibility signal. \TRACE{} is a research prototype; it was not assessed at the patient level against a reference standard (PCR or expert microscopy consensus), and it must not be used to inform clinical decisions. Any assertion regarding clinical utility requires prospective studies with external validation.

\subsection{Limitations and planned full-scale protocol}\label{sec:limits}

This pilot has several important limitations, each of which informs a component of the planned full-scale protocol:
\begin{enumerate}[leftmargin=2em,itemsep=2pt,topsep=2pt]
\item \textit{Evaluation design:} The pilot analysed 250 images from a single public dataset without external validation. Detector mAP@50 was calculated on the validation images and may therefore be overly optimistic. Performance at the species level under geographic domain shift was not assessed.
\item \textit{Confounding by confidence:} Predicted probability could plausibly confound the association between agreement and correctness. Section~\ref{sec:robust} examines added value beyond confidence, but the gate itself was not evaluated for calibration.
\item \textit{Thresholds and map geometry:} All thresholds were selected arbitrarily (Section~\ref{sec:xai}). Binarisation at the 75th percentile fixes the highlighted area at 25\% of crop pixels; consequently, annotation \IoU{} depends on the size of the annotated region relative to the tightly cropped image, with baseline and sensitivity analyses reported in Section~\ref{sec:robust}.
\item \textit{Explanation stability:} LIME used 100 perturbations rather than the commonly used default of 1,000, and run-to-run variability was not quantified. Accordingly, some disagreement between methods may result from explanation instability rather than model uncertainty.
\item \textit{Explanation ground truth:} Annotated bounding boxes serve as proxies for diagnostic regions of interest and do not represent the full morphological reasoning used by experts. The full-scale study will include parasitologist assessment of 10\% of explanation maps, with inter-rater $\kappa$ calculated.
\item \textit{Mixed-infection labels:} Results for mixed classes are based on 44 synthetically created cases and may not generalize to actual mixed infections; the sample size is limited. The full-scale study will use mixed infections confirmed by PCR.
\item \textit{Exploratory stratified analyses:} Comparisons between strata were not corrected for multiplicity, and several confidence intervals overlap.
\end{enumerate}

The planned full-scale protocol will use the Rwandan cohort (2,739 images), with PCR-corrected reference labels and image-level data partitioning; YOLOv12n-seg with a P2 head, under a pre-specified hypothesis that pv recall will improve by at least 5 points; the Lacuna Ghana and FASTMAL Nigeria cohorts for external pf validation and a South Asian \Pv-dominant cohort for \Pv; and hypotheses, thresholds, and analysis plans established before data access.

\subsection{Implications for explanation evaluation}

In medical imaging, explanation methods are frequently compared qualitatively by displaying heatmaps side by side. The modest agreement observed here (mean pairwise \IoU{} 0.58), together with the graded relationship between agreement and accuracy, indicates that quantifying agreement and reporting accuracy by method combination is more informative than visual inspection alone. It remains to be determined whether this finding holds on independent external data and provides value beyond predicted confidence.

\section{Conclusion}\label{sec:conclusion}

In this pilot feasibility study, agreement among Grad-CAM, LIME, and SHAP explanations could be quantified and was associated with the correctness of species-level malaria predictions: accuracy increased from 73.8\% to 89.5\% as the number of supporting methods rose from none to three, remained robust under image-level clustering, and provided modest discriminative value beyond predicted confidence alone. Morphological characteristics differed between pf and pv crops. These findings are preliminary, are based on a single public dataset, and do not demonstrate diagnostic performance or readiness for field use. Before explanation agreement can be regarded as a reliable trust signal, it must be validated on independent external data with PCR-confirmed mixed-infection labels and evaluated prospectively in clinical settings.

\section{Data and Code Availability}\label{sec:availability}

The Lacuna Malaria Datasets are publicly available on Kaggle (\url{https://www.kaggle.com/datasets/tandem03/lacuna-malaria-datasets})~\cite{lacuna2024}. The complete Google Colab notebook, including dataset download, detector and ensemble training, explanation generation, 14-feature extraction, stratified agreement analysis, and image-clustered evaluation, is provided as Supplementary File~1 (.ipynb). The Rwandan thick-smear cohort can be accessed under controlled conditions from the corresponding author after institutional ethics approval.

\newpage

\end{document}